\documentclass{article}

\usepackage[final]{nips_2016}
\usepackage{natbib}
\usepackage{hyperref}
\hypersetup{
  colorlinks   = true, 
  urlcolor     = blue, 
  linkcolor    = blue, 
  citecolor   = blue 
}

\usepackage[utf8]{inputenc} 
\usepackage[T1]{fontenc}    
\usepackage{hyperref}       
\usepackage{url}            
\usepackage{booktabs}       
\usepackage{amsfonts}       
\usepackage{nicefrac}       
\usepackage{microtype}      

\usepackage{lineno,hyperref}
\usepackage{graphicx, subfigure}			
\usepackage{amsmath}
\usepackage{amssymb}
\usepackage{mathtools}
\usepackage{amsmath}
\usepackage{multirow}

\usepackage{pdflscape}
\usepackage{afterpage}
\usepackage{capt-of}
\usepackage{lipsum}
\usepackage{geometry}
\usepackage[linesnumbered,ruled]{algorithm2e}

\title{Restricted Boltzmann machine to determine the input weights for extreme learning machines}

\author{
  Andre G. C. Pacheco \\
  Graduate Program in Computer Science, PPGI \\
  Federal University of Espírito Santo - UFES\\
  Vitória, ES - Brazil \\
  \texttt{pacheco.comp@gmail.com} \\
  \And
  Renato A. Krohling \\
  Production Engineering Department and Graduate Program in Computer Science, PPGI \\
  Federal University of Espírito Santo - UFES\\
  Vitória, ES - Brazil \\
  \texttt{krohling.renato@gmail.com} \\
  \And
  Carlos A. S. da Silva \\
  Graduate Program in Computer Science, PPGI \\
  Federal University of Espírito Santo - UFES\\
  Vitória, ES - Brazil \\
  \texttt{carlosalexandress@gmail.com} \\
}

\begin{document}

\maketitle

\begin{abstract}
The Extreme Learning Machine (ELM) is a single-hidden layer feedforward neural network (SLFN) learning algorithm that can learn effectively and quickly. The ELM training phase assigns the input weights and bias randomly and do not change them in the whole process. Although the network works well, the random weights in the input layer can make the algorithm less effective and impact on its performance. Therefore, we propose a new approach to determine the input weights and bias for the ELM using the restricted Boltzmann machine (RBM), which we call RBM-ELM. We compare our new approach with a well-known approach to improve the ELM and a state of the art algorithm to select the weights for the ELM. The results show that the RBM-ELM outperforms both methodologies and achieve a better performance than the ELM.
\end{abstract}

\section{Introduction}
The extreme learning machine (ELM) is an approach for training single-hidden layer feedforward neural network (SLFN) proposed by \cite{huang2004} and \citet{huang2006}. Its use has become popular due to its fast and analytically training phase. Comparing it with backpropagation \citet{rumelhart1988}, the most common algorithm used to train neural networks, its learning stage can be thousand of times faster and it can achieve a better generalization \citet{huang2006}.

The ELM learning algorithm assigns the weights from the input layer to hidden layer ($\mathbf{W}$) randomly. Then, it computes the weights from hidden layer to output layer ($\boldsymbol{\beta}$) analytically using the least-squares method. The ELM has been used to solve many different problems, such as classification data \citet{huang2010, xu2017}, time series forecasting \citet{butcher2013}, regression problem \citet{huang2012, shihabudheen2017}, among others.

Although \citet{huang2004} proved the universal approximation capability of SLFNs with unchanged random weights throughout the whole training phase, this issue has attracted the efforts of many researchers. As shown in \citet{wang2011}, the randomness of the input weights can make the algorithm less effective, depending on the assigned values for $\mathbf{W}$. Moreover, this point also influences the algorithm performance, i.e., the ELM output is not quite stable \citet{wang2015}. Due to this issue, many approaches have been proposed aiming to improve the generalization and performance of the ELM. Some of these approaches avoid assigning the input weights by providing a different way to compute the values of the hidden neurons, i.e., the feature map. One standard approach is the K-ELM, presented by \cite{huang2012}. It is a deterministic methodology that uses a suitable kernel function specified by the user to compute the hidden neurons. The K-ELM can achieve good performance, however, it does not work well when the database grows, because it demands a large computational time and memory consuming. We can highlight other methodologies, such as the PCA-ELM, a deterministic algorithm proposed by \cite{castano2013}, which is used to initiate any hidden neurons based on principal component analysis (PCA) and the PL-ELM, an architecture developed by \cite{henriquez2017}, which is based on a non-linear layer in parallel by another non-linear layer and with entries of independent weights. Deep learning techniques have also been used to improve the ELM performance. \cite{kasun2013} presented the ELM autoencoder (ELM-AE) and \cite{sun2017} expanded it developing the generalized ELM autoencoder (GELM-AE). Inspired by ELM-AE, \cite{tissera2016} proposed a deep neural network (DBN) using ELMs as a stack of supervised autoencoders. All these approaches are used with the same goal of the K-ELM and PCA-ELM, i.e., to extract the feature map of the first layer and to avoid the random weights assignment. 

Other approaches focus on finding a way to provide a better initialization of the input weights, and consequently improving the ELM performance. \cite{han2013} introduced an optimization of the input weights via particle swarm optimization (PSO). \cite{cervellera2016} presented an algorithm that replaces the random initialization by a deterministic one using low-discrepancy sequences (LDS). Recently, \cite{wang2017} proposed the ELM-RO, a state of the art approach that is mainly based on the Gram-Schmidt orthogonalization of the input weights, which achieves a better performance than the standard ELM and other algorithms that compute the weights for the ELM.

In this work, our main contribution is to present a new approach to determine the input weights for the ELM using the restricted Boltzmann machine (RBM) \citet{smolensky1986, hinton2002}, which we call RBM-ELM. Basically, the RBM is an energy-based system that can learn the probability distribution of a database by unsupervised learning. As autoencoders, RBMs are widely used to compose DBNs \citet{hinton2006}. Therefore, the architecture presented by \cite{tissera2016}, the stack of autoencoders to extract features from the data and set it as the ELM hidden nodes, can be done by using RBMs. Indeed, it was introduced by \cite{le2014}. Nonetheless, we need to make clear this is not our intention in this work. We just use one RBM to determine good values for the ELM input weights and then proceed with the standard ELM training. As we present throughout this paper, we need just a few RBM training epochs to set the ELM input weights. In the experiments section, we show that our approach has a relatively fast learning phase and achieves a good performance. Furthermore, we compare the RBM-ELM with the well-known ELM-AE and with the state of the art ELM-RO. Experimental results show that the RBM-ELM can outperform them for the benchmarks used in this paper. The remainder of the paper is organized as follows. In section \ref{sec:background}, we present a brief background on ELM and RBM.  Section \ref{sec:the-approach} describes our new approach. In section \ref{sec:results}, present the experimental results. Lastly, in section \ref{sec:conclusion} we draw our final conclusions.

\section{Background} \label{sec:background}
\subsection{Extreme learning machine} 
\begin{figure}[h]
\begin{center}
\includegraphics[scale=0.20]{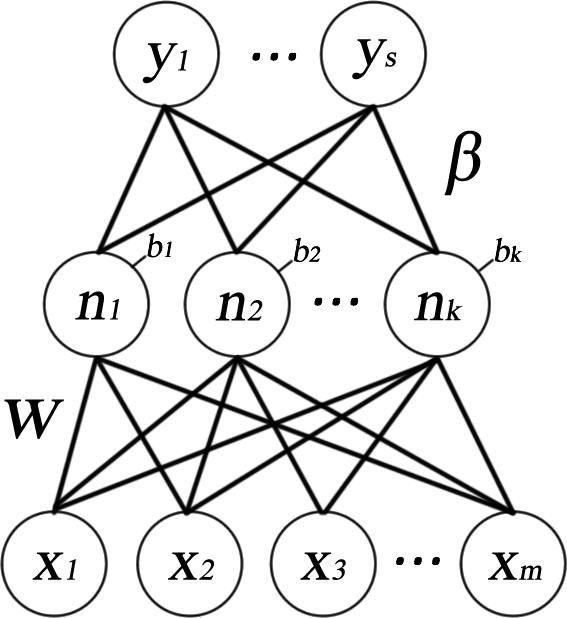}
\caption{The architecture of a single-hidden layer feedforward neural network}
\label{fig:SLFN}
\end{center}
\end{figure}

The extreme learning machine approach was developed specifically to handle with SLFN architecture. We depict in Figure \ref{fig:SLFN} an SLFN, where \textbf{x} is the input data, \textbf{y} is the output layer, \textbf{W} is the weights matrix of the input layer, $\boldsymbol{\beta}$ is the weights matrix of the hidden layer and $\mathbf{b}$ is the bias of the input layer. We represent all these notation as follows:  

\begin{equation}
	\label{eq:matrizesELM}
\mathbf{x} = [x_1, \cdots, x_m, 1]
\quad	
	\mathbf{W} = \begin{bmatrix}
 w_{11}& \cdots  & w_{1k} \\ 
\vdots & \ddots  & \vdots \\ 
 w_{m1}& \cdots  & w_{mk} \\
b_1 & \cdots & b_k
\end{bmatrix}
\quad
\boldsymbol{\beta} = \begin{bmatrix}
 \beta_{11}& \cdots  & \beta_{1s} \\ 
\vdots & \ddots  & \vdots \\ 
 \beta_{k1}& \cdots  & \beta_{ks}
\end{bmatrix}
\quad
\mathbf{y} = [y_1, \cdots , y_s]
\end{equation}

\noindent where $m$, $s$ and $k$ are the number of input, output and hidden neurons, respectively. Observe that the weights and bias are put together in the same matrix to ease the computation.

In the ELM algorithm, the matrix \textbf{W} is initialized randomly by sampling all the weight values from a continuous distribution, normally the uniform distribution in the interval [-1,1], and it does not change during the whole learning phase. \cite{huang2006} proved that we can obtain the matrix $\boldsymbol{\beta}$ performing an analytical process maintaining the universal approximation capability for the SLFNs. After sampling the weights we compute the hidden neurons, that is, the feature map (\textbf{H}), as follows:

\begin{equation}
	\label{eq:matrxzH}
\mathbf{h}^i = [x^i_1, \cdots, x^i_m, 1]
\times
\begin{bmatrix}
 w_{11}& \cdots  & w_{1k} \\ 
\vdots & \ddots  & \vdots \\ 
 w_{m1}& \cdots  & w_{mk} \\
b_1 & \cdots & b_k
\end{bmatrix}
\Rightarrow 
\mathbf{H} = \begin{bmatrix}
 f(\mathbf{h}^1)\\ 
 f(\mathbf{h}^2) \\
\vdots\\
 f(\mathbf{h}^N)
\end{bmatrix}_{N \times k}
\end{equation}

\noindent where $f(.)$ is the activation function, e.g., a logistic function, $i = (1, \hdots, N)$ and $N$ is the number of samples in the training set. Next, we compute $\boldsymbol{\beta}$ by solving the linear system through a simple generalized inverse operation, as described by:

\begin{equation}
	\label{eq:solveELM}
	\mathbf{H} \boldsymbol{\beta} = \mathbf{Y} \rightarrow \boldsymbol{\beta} = \mathbf{H}^\dagger \mathbf{Y}
\end{equation}

\noindent where $\mathbf{H}^\dagger$ is the Moore-Penrose generalized inverse of \textbf{H} \citep{serre2002}. The Moore-Penrose based solution is one of the least-square solutions for a general linear system. It can achieve: the smallest training error, the smallest norm of the weights and, as consequence, a good generalization performance. Moreover, it does not get stuck in local as the gradient descent-based learning methods \citep{huang2006}. A pseudocode of the ELM is presented in Algorithm \ref{alg:elm}. 

\begin{algorithm}[h]
	\label{alg:elm}
    \caption{Training an SLFN by the ELM procedure}
   	\textbf{Function} elm\_training $(\mathbf{X},\mathbf{Y},k, \textbf{W}, \textbf{b})$: \\
    \SetKwInOut{Input}{Input}
    \SetKwInOut{Return}{Return}
    \Input{A training set $\{\mathbf{X}, \mathbf{Y}\}$; \\ The number of hidden neurons $k$; \\ The input weights \textbf{W} and the bias \textbf{b}; \\}

	\If{\textbf{W} and \textbf{b} are \textit{empty}}{
		Sampled them by a uniform distribution in the interval $[-1,1]$;
	}
	
	Compute the feature map \textbf{H} from equation \ref{eq:matrxzH};\\
	Compute $\boldsymbol{\beta}$ from equation \ref{eq:solveELM};\\

    \Return {\textbf{W}, \textbf{H} and $\boldsymbol{\beta}$;}
\end{algorithm}

\subsection{Restricted Boltzmann machine} \label{sec:rbm}
\begin{figure}[h]
\begin{center}
\includegraphics[scale=0.25]{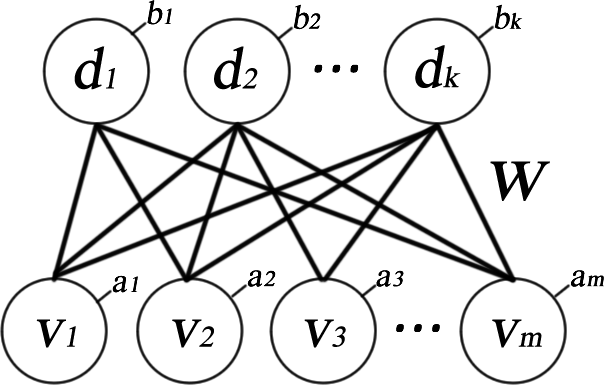}
\caption{The architecture of a RBM with $m$ visible nodes and $k$ hidden nodes}
\label{fig:rbm}
\end{center}
\end{figure}

The restricted Boltzmann machine (RBM) is a stochastic network composed of a visible layer (\textbf{v}) and a hidden layer (\textbf{d)}. As illustrated in Figure \ref{fig:rbm}, there is no connection within a layer, \textbf{v} and \textbf{d} have symmetric connectivity \textbf{W} and each layer has its own bias, \textbf{a} and \textbf{b}. During the training phase, the RBM learns the probability distribution over the input data training through unsupervised learning. Therefore, this network has been applied in different tasks, such as feature extraction, pattern recognition, dimensionality reduction, data classification etc \citep{hinton2010}.  

\noindent Originally, the RBM was proposed for binary input data. However, \cite{hinton2006Sal} extended it for continuous input data. In this paper we use the continuous approach.

In the RBM model, each configuration (\textbf{v},\textbf{d}) has an associated energy value defined by: 

\begin{equation}
	\label{eq:rbmEnergy}
	E(\mathbf{v},\mathbf{d}; \boldsymbol{\theta}) = -\sum_{i=1}^{m} \frac{(v_{i}-a_{i})^2}{2\sigma_i^2} - \sum_{j=1}^{k}b_{j}d_{j} - \sum_{i,j=1}^{m,k} \frac{v_{i}}{\sigma^2} d_{j}w_{ij}  
\end{equation}

\noindent where $\boldsymbol{\theta} = (\mathbf{W}, \mathbf{a}, \mathbf{b})$. The joint probability of (\textbf{v}, \textbf{d}) is computed as follows:

\begin{equation}
\label{eq:joinProbRBM}
	p(\mathbf{v},\mathbf{d}; \boldsymbol{\theta}) = \dfrac{e^{-E(\mathbf{v},\mathbf{d}; \boldsymbol{\theta})}}{\sum_{\mathbf{v},\mathbf{d}}e^{-E(\mathbf{v},\mathbf{d}; \boldsymbol{\theta})}} 
\end{equation}

In general, the goal of the RBM learning algorithm is estimating $\boldsymbol{\theta}$ that decreases the energy function \citep{hinton2010}. \cite{hinton2002} proposed an efficient algorithm for training RBMs, known as contrastive divergence (CD). The CD is an unsupervised algorithm that uses an iterative process known as Gibbs sampling. The main idea of this algorithm is initializing the visible layer with training data and then perform the Gibbs sampling. The CD is an easy and fast learning algorithm. Therefore, the most important use of the RBM is as learning modules that are composed to form deep belief nets \citep{hinton2010}.

In order to perform the CD algorithm, we need to compute $p(\mathbf{d}|\mathbf{v})$ as follows:

\begin{equation}
	\label{eq:h-given-v}
	p(d_{j}=1|\mathbf{v} ; \boldsymbol{\theta}) = \phi (b_{j}+\sum_{i=1}^{m} v_{i}w_{ij})
\end{equation}

\noindent where $\phi (x) = \frac{1}{1+e^{-x}}$, the logistic function. Then, we compute $p(\mathbf{v}|\mathbf{d})$ by:

\begin{equation}
	\label{eq:v-given-h}
	p(v_{i}=v|\mathbf{d} ; \boldsymbol{\theta}) = N (v | a_{i}+ \sum_{j=1}^{k} d_{j}w_{ij},\sigma^{2})
\end{equation}

\noindent where $N$ is the normal distribution with mean $v$ and standard deviation $\sigma^2$. 

\begin{figure}[h]
\begin{center}
\includegraphics[scale=0.18]{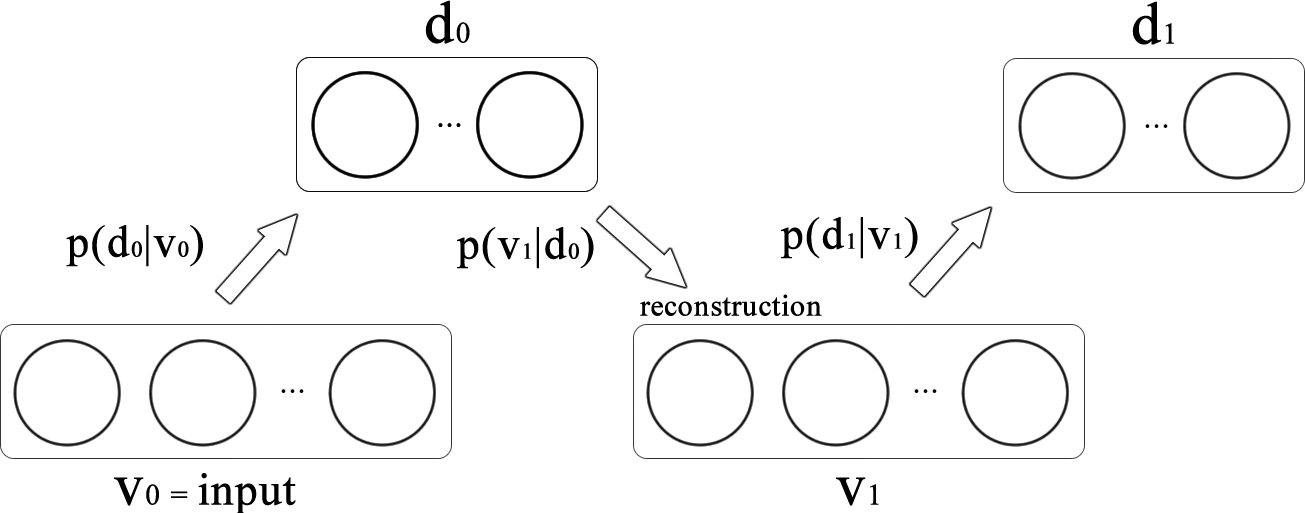}
\caption{The Gibbs sampling procedure on CD algorithm}
\label{fig:cd}
\end{center}
\end{figure}

The Gibbs sampling procedure is illustrated in Figure \ref{fig:cd}. As we can note, first we initialize the visible layers with the training data. Next, we estimate $\mathbf{d}_0$ using the equation \ref{eq:h-given-v}. From $\mathbf{d}_0$ we estimate $\mathbf{v}_1$ by equation \ref{eq:v-given-h}. This step is called reconstruction. Finally, from $\mathbf{v}_1$ we estimate $\mathbf{d}_1$ using equation \ref{eq:h-given-v} again. This whole procedure can be done $z$ times, however, with just one iteration the algorithm works quite well \citep{hinton2010}. After proceeding with the Gibbs sampling, the CD update rules for $\boldsymbol{\theta}$ are described by the following equations:

\begin{equation} \label{eq:dw}	
	\mathbf{W^{t+1}} = \mathbf{W^{t}} +	\Delta \mathbf{W^{t}} \\ \rightarrow  \Delta \mathbf{W^{t}} = {\eta} (\mathbf{v_0}\mathbf{d_0^T} - \mathbf{v_1}\mathbf{d_1^T}) - {\rho} \mathbf{W^{t}} + {\alpha} \Delta \mathbf{W^{t-1}}	
\end{equation}		
	
\begin{equation} \label{eq:da}	
		\mathbf{a^{t+1}} = \mathbf{a^{t}} + \Delta \mathbf{a^{t}} \rightarrow \Delta \mathbf{a^{t}} = {\eta} (\mathbf{v_{0}} - \mathbf{v_{1}}) + {\alpha} \Delta \mathbf{a^{t-1}}
\end{equation}	
	
\begin{equation} \label{eq:db}	
		\mathbf{b^{t+1}} = \mathbf{b^{t}} + \Delta \mathbf{b^{t}} \rightarrow \Delta \mathbf{b^{t}} = {\eta} (\mathbf{d_{0}} - \mathbf{d_{1}}) + {\alpha} \Delta \mathbf{b^{t-1}}
\end{equation}

\noindent where ${\eta}$, ${\rho }$ and ${\alpha}$ are known as learning rate, weight decay and momentum, respectively. \cite{hinton2010} suggests ${\eta} = 0.01$, ${\rho } = [0.01, 0.0001]$ and ${\alpha} = 0.5$ for the first five iterations and ${\alpha} = 0.9$ otherwise. Usually, the terms of $\boldsymbol{\theta}$ are initialized randomly. Further, \cite{hinton2010} also suggests to divide the training set into small mini-batches of 10 to 100 cases. A pseudocode of the constrastive divergence is presented in Algorithm \ref{alg:cd}.

\begin{algorithm}[H]
	\label{alg:cd}
    \caption{The contrastive divergence procedure}
    \SetKwInOut{Input}{Input}
    \SetKwInOut{Return}{Return}
	\textbf{Function} rbm\_training $(\mathbf{X},k,\eta,\rho,\alpha, bs, it)$: \\
    \Input{The training data \textbf{X} with $N$ samples;\\ The number of hidden neurons $k$; \\ The parameters ${\eta}$, ${\rho }$ and ${\alpha}$; \\ The batch size $bs$;\\ The maximum number of iterations $it$\\ }

	Sample $\boldsymbol{\theta}$ from a normal distribution with $\mu=0.1$ and $\sigma=1$; \\
	Split \textbf{X} in batches $\mathbf{X_b}$ with \textit{bs} samples; \\
	Initialize $i = j = 0$; \\
	\While{$j < it$ or $\boldsymbol{\theta}$ converged}{		
		\For{each batch $\mathbf{X_b}$}{
			\While{$i < bs$}{
				$\mathbf{v}_0$ = $\mathbf{X_b}_i$
				Sample $\mathbf{d}_0$ from equation \ref{eq:h-given-v}; \\
				Sample $\mathbf{v}_1$ from equation \ref{eq:v-given-h}; \\
				Sample $\mathbf{d}_1$ from equation \ref{eq:h-given-v}; \\
				Update $\boldsymbol{\theta}$ from equations \ref{eq:dw}, \ref{eq:da} and \ref{eq:db};\\
				$i = i + 1$
			}		
		}
		$j = j + 1$
	}
    \Return {$\boldsymbol{\theta}$;}
\end{algorithm}

\section{A new approach to determine weights for ELMs} \label{sec:the-approach}
\begin{figure}[h]
\begin{center}
\includegraphics[scale=0.20]{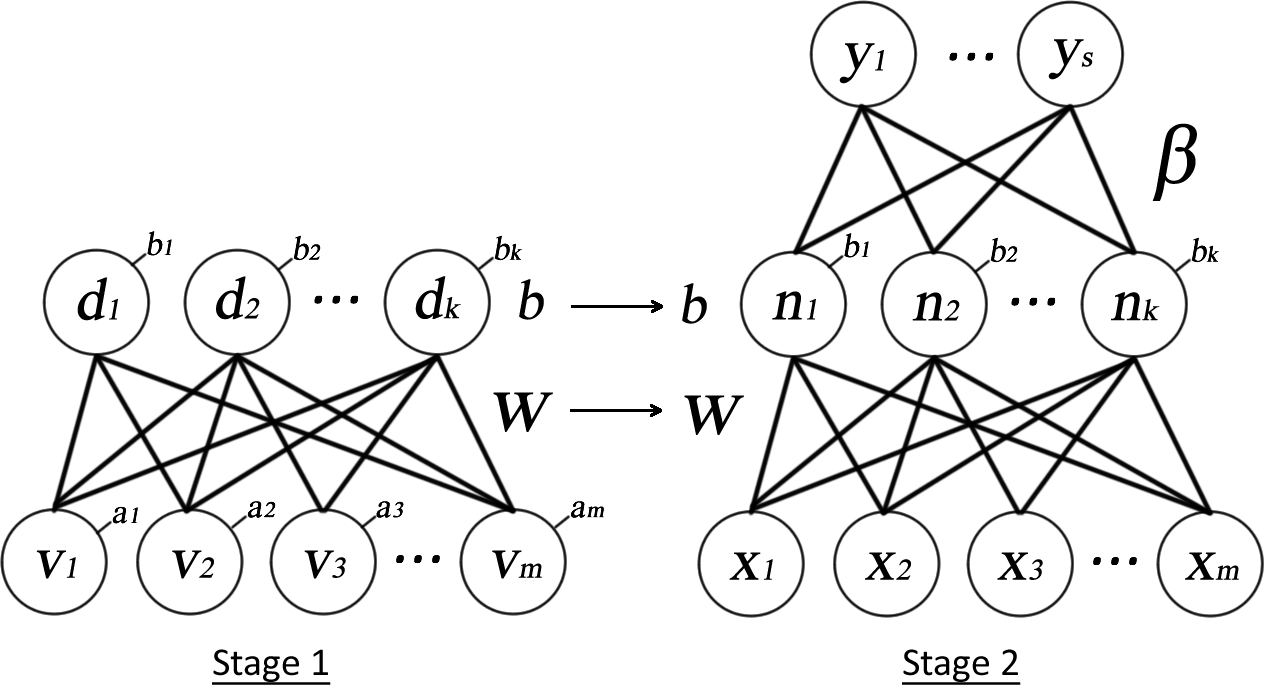}
\caption{An illustration of the RBM-ELM approach}
\label{fig:rbm-elm}
\end{center}
\end{figure}

In this section, we introduce the new approach to determine weights for the extreme learning machine using the restricted Boltzmann machine, which we call RBM-ELM. Our focus is on the input weights \textbf{W}. Since the ELM assigns \textbf{W} at random and it is used to compute \textbf{H} and $\boldsymbol{\beta}$, there inevitably exists a set of nonoptimal input weights and hidden biases values, which may influence on the ELM performance \citep{han2013}. Thus, the RBM-ELM main idea is to replace the ELM input weights and bias by the RBM visible weights and hidden bias, as shown in Figure \ref{fig:rbm-elm}.

As we can note in Figure \ref{fig:rbm-elm}, the RBM-ELM approach has two stages. In the first stage, we compute the input weights and bias for the ELM through the RBM training. In the second stage, the results of the first stage are used to set \{\textbf{W}, \textbf{b}\} and then to proceed with the standard ELM training. To better describe these two stages, consider a training set composed by $\{\mathbf{X}_t, \mathbf{Y}_t\}$, where $\mathbf{X}_t$ and $\mathbf{Y}_t$ are the input and output training data, respectively.  First, the RBM is trained with $\mathbf{X}_t$. All the knowledge obtained by this network is stored on its weights and bias. So, after the RBM training, we set the ELM input weights and bias with the same values of the RBM weights connections and hidden bias, respectively. Next, we carry out the ELM training using $\mathbf{X}_t$, $\mathbf{Y}_t$ and the computed \{\textbf{W}, \textbf{b}\} to compute \textbf{H} and $\boldsymbol{\beta}$. Since we use the same input training data to feed the visible and input layer in the RBM and ELM, respectively, \textbf{v} and \textbf{x} have the same shape. Thus, to guarantee the same shape for \textbf{W} and \textbf{b} on both networks, we need to set the same number of hidden neurons for both algorithms. A pseudocode of this approach is described in Algorithm \ref{alg:rbm-elm}.

\begin{algorithm} \label{alg:rbm-elm}
    \SetKwInOut{Input}{Input}
    \SetKwInOut{Return}{Return}
    \textbf{Function} rbm\_elm\_training $(\{\mathbf{X}_t, \mathbf{Y}_t\},k,\eta,\rho,\alpha)$: \\

    \Input{The training set $\{\mathbf{X}_t, \mathbf{Y}_t\}$ with $N$ samples; \\ The number of neurons $k$; \\ The parameters ${\eta}$, ${\rho }$ and ${\alpha}$;}
    From Algorithm \ref{alg:cd} call rbm\_training $(\mathbf{X}_t,k,\eta,\rho,\alpha)$ to get \textbf{W} and \textbf{b}; \\ 
	From Algorithm \ref{alg:elm} call elm\_training $(\mathbf{X}_t,\mathbf{Y}_t,k, \textbf{W}, \textbf{b})$ to get \textbf{H} and $\boldsymbol{\beta}$;\\
	
	\Return{\textbf{W}, \textbf{H} and $\boldsymbol{\beta}$;}
    \caption{The RBM-ELM procedure}
\end{algorithm}

As mentioned earlier, a stack of RBMs has been used to improve the ELM performance \citep{le2014}. Nonetheless, its goal is to find the feature map for the input layer, i.e., setting the matrix \textbf{H} on the ELM. As a drawback, this method may take a long computational time to achieve a good feature map, which removes from the ELM one of its great advantages: the fast training phase. Our approach aims to take advantage of the RBM generalization capability. Since the RBM models the probability distribution over the data training, when we compute the ELM weights and bias from an RBM, we also transfer the knowledge obtained by the RBM training phase to the ELM. Consequently, we improve the feature map extracted by the input layer in the ELM.  

As we can note in Algorithm \ref{alg:cd}, the RBM training is carried out until the weights convergence or a specific number of epochs. Comparing with the ELM training phase, the RBM training may take a considerable computational time if we wait for the convergence or set a high value for the number of epochs. However, the reconstruction error on the entire training set falls rapidly and consistently at the start of learning and then more slowly \citep{hinton2010}. So, the weights converge nearly to their final values after only a few epochs, but the further fine tuning takes much longer \citep{yosinski2012}. As our intention is to determine the input weights and bias for the ELM, the proposed approach works well performing just a few epochs of the RBM training. So, the RBM training phase does not affect too much the time-consuming of the whole algorithm. In the next section, we show the effect of the number of epochs on the algorithm performance.

\section{Experimental results} \label{sec:results}
In this section, we carry out two experiments for classification problems. First, we present a thorough example to better describe the RBM-ELM. Next, we compare our approach with the standard ELM, the state of the art ELM-RO, and the ELM-AE, using several well-known benchmarks. All procedures were implemented in Python and Tensorflow, and performed on an \textit{intel core i7-6} CPU @ 2.50 GHz PC with 8 GB of RAM and a Nvidia Geforce 940M. The code developed is available upon request.

\subsection{Illustrative example} \label{sec:case-study}
\begin{figure}[h]
\begin{center}
\includegraphics[scale=0.65]{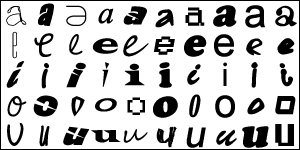}
\caption{Some samples from the vowels database}
\label{fig:vowels}
\end{center}
\end{figure}

In this illustrative example, we investigate the RBM-ELM configurations in order to improve the algorithm performance. We developed a vowels database, which contains 1380 samples with 276 examples for each vowel. In Figure \ref{fig:vowels} is shown some samples from the database, where each vowel is represented by a image with 30 $\times$ 30 pixels. Thus, the vowels database has 900 input features and five classification labels. 

\noindent The vowels database was shuffled and split to 70\% for training and 30\% for tests. The ELM and RBM-ELM were run 30 times to compute statistics and we use the mean and standard deviation of the classification accuracy as performance metric. In general, our goal in this section is to present a discussion about the RBM parameters, the number of epochs used to train a RBM, the running time of the whole approach, the number of hidden neurons, and compare the performance between the ELM and RBM-ELM. 

First of all, we need to choose the number of hidden neurons ($k$). We perform the ELM and choose $k$ empirically based on the best performance of this algorithm. To be fair, we decided to use the same value of $k$ to compare both methodologies. In table \ref{tab:elm-vowels} is described the ELM performance varying the number of hidden neurons. As we can note, the best performance is achieved when $k = 400$. Thus, we use this number of hidden neurons to perform the RBM-ELM as well.

\begin{table}[h]

\begin{center}
\caption{The ELM performance for vowels database varying the number of hidden neurons}
\begin{tabular}{c|c|c}\hline

\textbf{\# of neurons}& \textbf{Accuracy (\%)} & \textbf{Time (sec)} \\ \hline

100 & $79.677 \pm 1.985$ & $0.050 \pm 0.006$   \\

200 & $85.789 \pm 1.356$ & $0.096 \pm 0.010$  \\

300 & $86.771 \pm 1.555$ & $0.166. \pm 0.016$ \\

\textbf{400} & $ \mathbf{87.975} \pm \mathbf{1.500}$ & $\mathbf{0.239} \pm \mathbf{0.029}$    \\

500 & $86.972 \pm 1.721$ & $0.329 \pm 0.027$ \\

600 & $85.668 \pm 1.793$ & $0.436 \pm 0.026$ \\ \hline
\end{tabular}
\label{tab:elm-vowels}
\end{center}
\end{table}

As described in section \ref{sec:the-approach}, in order to apply the RBM-ELM, first we need to perform the stage 1 of the algorithm, that is, to train the RBM and compute the input weights and bias for the ELM. As we can note in Algorithm \ref{alg:rbm-elm}, the RBM has seven parameters, where $\mathbf{X}$ is the input data and \textit{k} is the number of hidden neurons, which is the same value that we found for ELM, $k=400$. For $\eta$, $\rho$, $\alpha$ and $bs$, we use the values in the interval suggested by \cite{hinton2010}, presented in section \ref{sec:rbm}. These values are described in Table \ref{tab:rbm-elm-vowels} and their fine tune was achieved empirically by tests. Finally, we need to choose the number of maximum iterations $it$, which affect directly the time consuming of the algorithm. Indeed, we could wait for $\boldsymbol{\theta}$ convergence, however it may take a long time and we do not need to do that. As described in section \ref{sec:the-approach}, we need a few iterations to put the weights nearly to their final values and these values are good enough to improve the ELM. After finishing the stage 1, we perform the stage 2 of the algorithm with the weights and bias computed from the stage 1. In table \ref{tab:rbm-elm-vowels} is described the RBM-ELM performance varying \textit{it} and computing the time consuming for each scenario.

\begin{table}[h]

\begin{center}
\caption{The RBM-ELM performance for vowels database varying the iteration number of the RBM training}
\begin{tabular}{c|c|c}\hline

\multicolumn{3}{c}{\tiny \textbf{RBM parameters:} $k=400$, $\eta = 0.001$, $\rho=0.01$, $\alpha=0.5/0.9$ and $bs=100$} \\ \hline

\textbf{\# of iterations (\textit{it})}& \textbf{Accuracy (\%)} & \textbf{Time (sec)} \\ \hline
10  & $90.708 \pm 1.518$  & $1.421 \pm 0.115$    \\

30  & $90.442 \pm 1.144$  & $3.617 \pm 0.120$   \\

\textbf{50}  & $\mathbf{92.769} \pm \mathbf{1.078}$  & $\mathbf{6.008} \pm \mathbf{0.099}$  \\

100 & $90.724 \pm 1.303$  & $11.667 \pm 0.096$    \\

300 & $90.611 \pm 1.420$  & $33.521 \pm 0.311$   \\

500 & $89.067 \pm 1.441$  & $54.901 \pm 1.336$  \\ \hline

\end{tabular}

\label{tab:rbm-elm-vowels}
\end{center}
\end{table}

\noindent As we can note in Table \ref{tab:rbm-elm-vowels}, the RBM can improve the ELM performance even with a low number of iterations. For $it=50$ the algorithm achieved the best performance. Nonetheless, for \textit{it} less than $100$, the approach may get good performance. However, the higher is the value of $it$, the higher is the running time. Comparing the RBM-ELM with $it=50$ and the best ELM performance in Table \ref{tab:elm-vowels}, our approach improve the classification accuracy in almost 5\% and decrease the standard deviation around 0.5\%. On the other hand, the standard ELM training is 25 times faster than RBM-ELM with $50$ iterations. In fact, we need to make a trade-off between accuracy and computational time. In this case, the improved performance is very desired.

\cite{huang2006} affirm that the ELM learning not only tends to reach the smallest training error but also the smallest norm of weights. \cite{bartlett1998} states that the smaller the norm of weights, the better generalization performance a feedforward network tends to have. So, we also compare the norm of the input weights obtained by the ELM and RBM-ELM for the vowels database. Based on the best performance of both ELM and RBM-ELM in Table \ref{tab:elm-vowels} and \ref{tab:rbm-elm-vowels}, respectively, the norm for the ELM is $346.602$ and for RBM-ELM is $2.839$. Hence, according to \cite{huang2006} and  \cite{bartlett1998}, for this database, the RBM-ELM tends to have a better generalization performance than the ELM.

\begin{figure}[h]
\begin{center}
\includegraphics[scale=0.15]{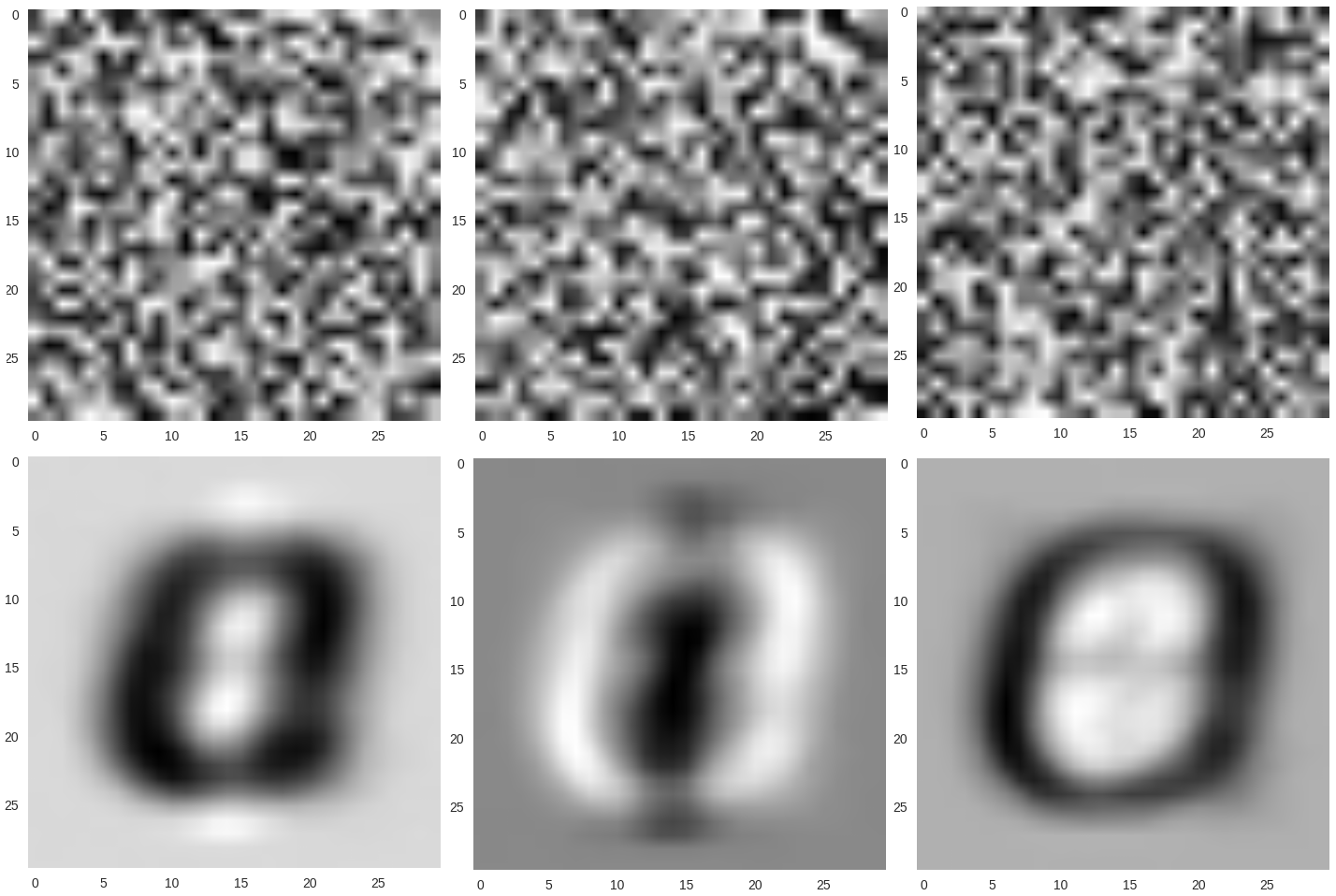}
\caption{The plot of the input weights for the ELM (above) and RBM-ELM (below).}
\label{fig:w-plot}
\end{center}
\end{figure}

In Figure \ref{fig:w-plot} is illustrated the graphic difference between the ELM and RBM-ELM input weights. Since the ELM set the weights randomly, its plot looks like a noise. On the other hand, in the RBM-ELM the input weights show some low-level features, even using 50 iterations of the RBM training phase. These plots help to explain why RBM-ELM achieves a better performance than ELM in this database.

\subsection{Standard benchmarks} \label{sec:benchs}
In order to evaluate the performance of the RBM-ELM we carry out experiments with different well-known classification benchmarks from UCI repository \citep{UCI}. We still compare the RBM-ELM performance with the standard ELM. In addition, we include two more algorithms, the ELM-RO and ELM-AE, both discussed in the introduction. All the databases used in this experiment is described in Table \ref{tab:databases}. In this table, each database with permutation equal to \textit{yes} was shuffled and split to 70\% for training and 30\% for tests. For permutation equal to \textit{no}, it means that this database has a test partition. The four algorithms were run 30 times to compute statistics. We used the mean and standard deviation of the classification accuracy as the performance metric. Further, we perform the non-parametric Friedman test followed by Wilcoxon test as a pos hoc to compare the algorithms performance \citep{derrac2011}.

\begin{table}[h]
\begin{center}
\caption{The classifications databases used in this experiment}
\begin{tabular}{c|c|c|c|c}\hline

\textbf{Database}& \textbf{\# of samples}&   \textbf{\# of features} &\textbf{\# of labels} & \textbf{Permutation} \\ \hline

\textit{Alphabet}          &11960  &900   &26  & Yes\\

\textit{Credit Australia}  &690    &14    &2  & Yes\\

\textit{Diabetic}          &1151   &19    &2  & Yes \\

\textit{DNA}               &3186   &180   &3  & No\\

\textit{Gisette}           &7000   &5000  &3  & No   \\

\textit{Isolet}            &7797   &617   &26 & No\\

\textit{Madelon}  	       &2600   &500   &2  & Yes\\

\textit{Spam}              &4601   &57    &2  & Yes\\

\textit{Urban land cover}  &675    &147   &9  & Yes\\ \hline

\end{tabular}

\label{tab:databases}
\end{center}
\end{table}

In Table \ref{tab:rbm-elm-config} is described the RBM-ELM parameters configuration for each database presented in Table \ref{tab:databases}. These parameters were selected empirically using the same process detailed in section \ref{sec:case-study}. For a fair comparison, the ELM-RBM, ELM-AE, and ELM-RO use the same number of hidden neurons of the ELM, which is described as $k$ in Table \ref{tab:rbm-elm-config}. In Table \ref{tab:results} is reported the performance of all algorithms for each database used in this experiment. In this table is described the mean and standard deviation of the classification accuracy and the mean value of the running time for each approach. Moreover, in order to improve the results visualization, in Figure \ref{fig:boxplots} is depicted the boxplots for all approaches also for each database. 

\begin{table}[h]

\begin{center}
\caption{The RBM-ELM parameters configuration for each database}
\begin{tabular}{c|c}\hline

\textbf{Database}& \textbf{RBM-ELM parameters} \\ \hline
\textit{Alphabet}     & $k=950$, $it = 25$, $\eta = 0.001$, $\rho=0.001$, $\alpha=0.5/0.9$ and $bs=150$  \\
\textit{Credit Australia}   & $k=35$, $it = 25$, $\eta = 0.1$, $\rho=0.1$, $\alpha=0.5/0.9$ and $bs=150$  \\
\textit{Diabetic}     & $k=50$, $it = 50$, $\eta = 0.01$, $\rho=0.001$, $\alpha=0.5/0.9$ and $bs=100$ \\
\textit{DNA}          & $k=250$, $it = 50$, $\eta = 0.0001$, $\rho=0.01$, $\alpha=0.5/0.9$ and $bs=250$\\
\textit{Gisette}      & $k=850$, $it = 50$, $\eta = 0.001$, $\rho=0.01$, $\alpha=0.5/0.9$ and $bs=250$ \\
\textit{Isolet}       & $k=1250$, $it = 25$, $\eta = 0.01$, $\rho=0.01$, $\alpha=0.5/0.9$ and $bs=250$\\
\textit{Madelon}      & $k=250$, $it = 50$, $\eta = 0.001$, $\rho=0.001$, $\alpha=0.5/0.9$ and $bs=50$\\
\textit{Spam}         & $k=150$, $it = 50$, $\eta = 0.001$, $\rho=0.0001$, $\alpha=0.5/0.9$ and $bs=100$\\
\textit{Urban land cover}        & $k=200$, $it = 25$, $\eta = 0.01$, $\rho=0.01$, $\alpha=0.5/0.9$ and $bs=20$\\ \hline

\end{tabular}

\label{tab:rbm-elm-config}
\end{center}
\end{table}

According to the results presented in the tables and boxplots, we can note that the ELM-AE, ELM-RO and RBM-ELM have better performance than ELM for most of databases. Only for \textit{credit Australia} and \textit{spam} databases, these approaches were not able to improve the classification performance. In fact, for these two databases, all four algorithms got the same performance. However, for the remaining databases, the ELM was improved by at least one of the others methods. So, the ELM has the lowest overall accuracy of all methods. On the other hand, the ELM is still the fastest algorithm among all. Indeed, this is expected since the other approaches require more processing time to compute the input weights. When we look at the RBM-ELM performance, we can note that it achieves the best performanc, alone or followed by another method for all database. As consequence, the RBM-ELM obtained the highest overall accuracy. However, it also gets the highest total time among all algorithms.

\afterpage{
\begin{landscape}
\begin{table}[]

\begin{center}
\caption{Algorithms performance for each database. In bold, the best algorithm(s) according to the statistical test}
\begin{tabular}{c|cc|cc|cc|cc}
\hline
\multirow{2}{*}{\textbf{Database}} & \multicolumn{2}{c|}{\textbf{ELM}} & \multicolumn{2}{c|}{\textbf{ELM-AE}} & \multicolumn{2}{c|}{\textbf{ELM-RO}}         & \multicolumn{2}{c}{\textbf{RBM-ELM}}  \\ \cline{2-9} 
& \textit{Accuracy (\%)} & \textit{Time (sec)} & \textit{Accuracy (\%)} & \textit{Time (sec)} & \textit{Accuracy (\%)} & \textit{Time (sec)} & \textit{Accuracy (\%)} & \textit{Time (sec)} \\ \hline

\textbf{Alphabet}     
&$72.784 \pm 0.713$ &$3.03$ &$76.934 \pm 0.777$ &$12.22$ &$73.285 \pm	0.655$ &$3.21$ &$\mathbf{78.677} \pm \mathbf{0.611}$ &$38.74$ \\

\textbf{Credit Australia}
&$85.732 \pm 2.292$ &$0.004$   &$86.025 \pm 2.237$ &$0.078$ &$85.829 \pm 2.280$ &$0.005$ &$86.070 \pm 1.960$ &$0.380$ \\

\textbf{Diabetic}
&$74.415 \pm 2.562$ &$0.010$   &$72.726 \pm 1.280$ &$0.016$   &$\mathbf{75.033} \pm \mathbf{1.866}$ &$0.011$   &$\mathbf{75.323} \pm \mathbf{1.996}$ &$0.442$ \\

\textbf{DNA}
&$89.232 \pm 0.827$ &$0.146$   &$93.931 \pm 0.549$ &$0.225$   &$94.139 \pm 0.364$ &$0.152$   &$\mathbf{94.592} \pm \mathbf{0.028}$ &$3.279$ \\

\textbf{Gisette}
&$92.130 \pm 0.809$ &$2.922$   &$\mathbf{96.383} \pm \mathbf{0.479}$ &$11.804$  &$94.953 \pm 0.519$ &$38.587$  &$\mathbf{96.116} \pm \mathbf{0.459}$ &$160.23$ \\

\textbf{Isolet}
&$94.032 \pm 0.385$ &$3.738$   &$\mathbf{95.244} \pm \mathbf{0.233}$ &$6.155$   &$94.746 \pm 0.298$ &$3.831$   &$\mathbf{95.135} \pm \mathbf{0.218}$ &$23.342$ \\

\textbf{Madelon}
&$55.393 \pm 1.732$ &$0.129$   &$57.234 \pm 1.429$ &$0.253$   &$66.145 \pm 1.421$ &$0.955$   &$\mathbf{82.286} \pm \mathbf{1.139}$ &$9.706$ \\ 

\textbf{Spam}
&$91.178 \pm 0.899$ &$0.096$   &$91.066 \pm 0.684$ &$0.175$   &$90.826 \pm 0.692$ &$0.099$   &$91.137 \pm 0.696$ &$1.715$ \\

\textbf{Urban land cover}
&$76.288 \pm 2.860$ &$0.044$   &$\mathbf{78.998} \pm \mathbf{3.080}$ &$0.119$   &$77.602 \pm 2.523$ &$0.045$   &$\mathbf{80.098} \pm \mathbf{2.589}$ &$2.161$ \\ \hline

\textbf{Overall mean/Total time}
&$81.243 \pm 1.453$ &10.119   &$83.171 \pm 1.194$ &31.045   &$83.618 \pm 1.179$ &46.895   &$\mathbf{86.604} \pm \mathbf{1.077}$ &239.995  \\ \hline  

\end{tabular}

\label{tab:results}
\end{center}
\end{table}

%
\begin{figure}[htb]
\centering
\subfigure[Alphabet]{\label{fig:a}\includegraphics[scale=0.41]{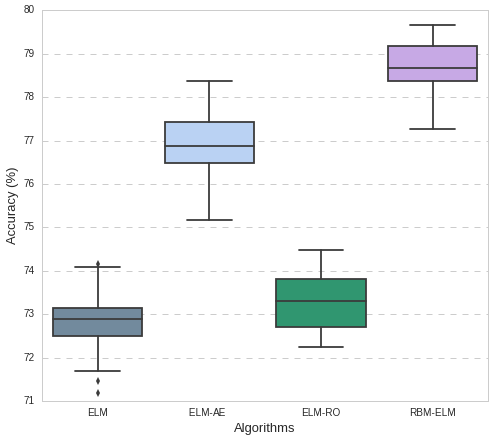}}
\subfigure[Credit Australia]{\label{fig:b}\includegraphics[scale=0.41]{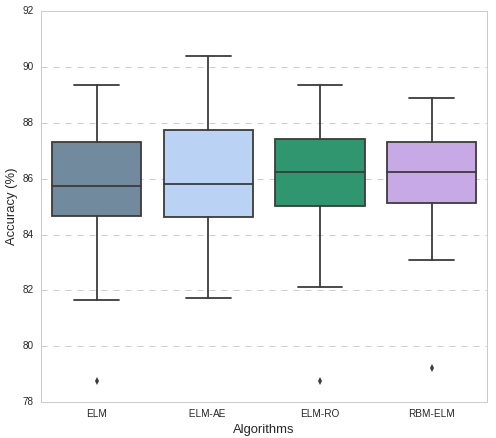}}
\subfigure[Diabetic]{\label{fig:c}\includegraphics[scale=0.41]{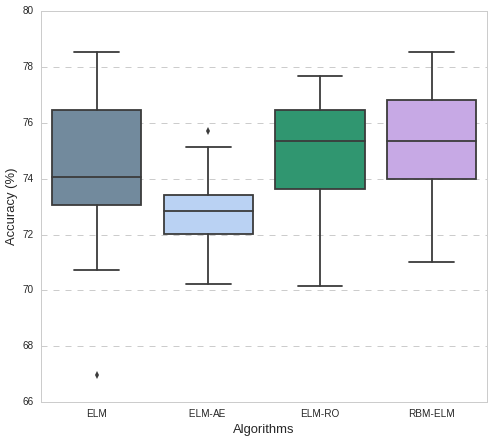}}
\subfigure[DNA]{\label{fig:d}\includegraphics[scale=0.41]{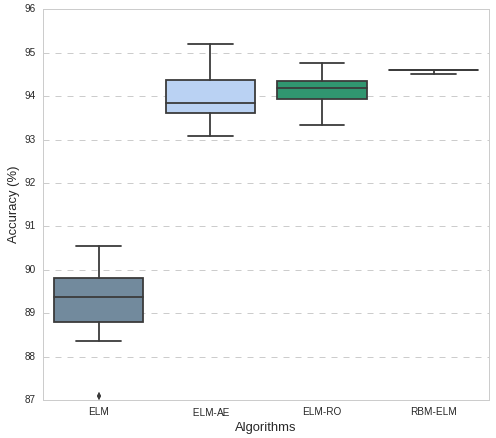}}
\subfigure[Gisette]{\label{fig:e}\includegraphics[scale=0.41]{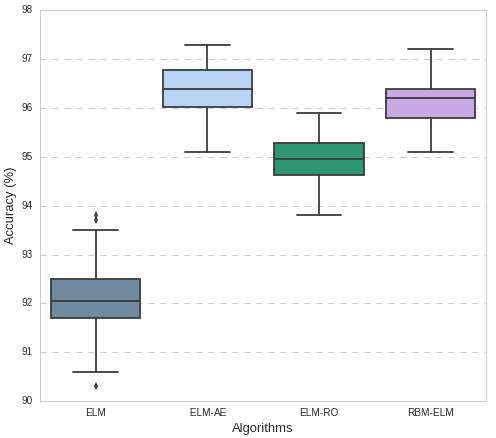}}

\end{figure}
\begin{figure}[htb] 
\centering

\subfigure[Isolet]{\label{fig:f}\includegraphics[scale=0.41]{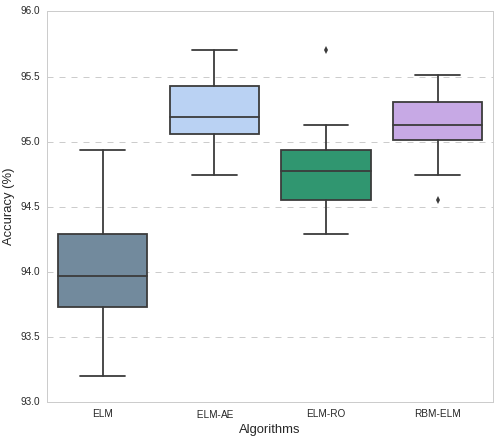}}
\subfigure[Madelon]{\label{fig:g}\includegraphics[scale=0.41]{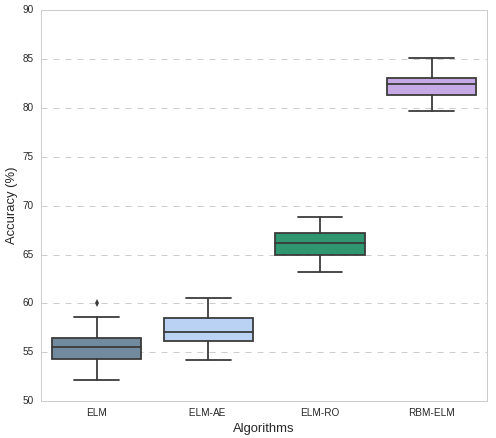}}
\subfigure[Spam]{\label{fig:g}\includegraphics[scale=0.41]{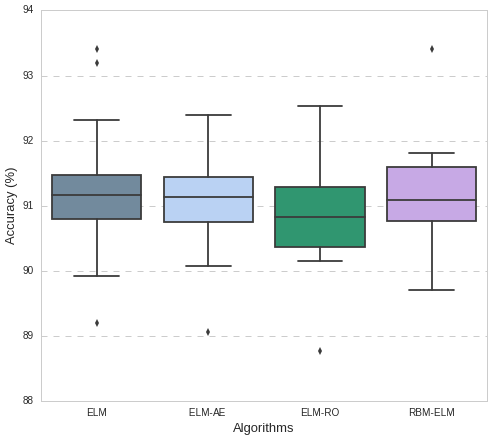}}
\subfigure[Urban land cover]{\label{fig:g}\includegraphics[scale=0.41]{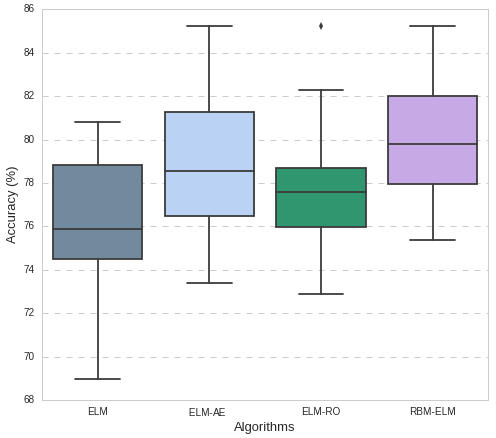}}

\caption{Boxplots for the algorithms performance for each database}
\label{fig:boxplots}
\end{figure}

   \end{landscape}    
    
}

We perform the Friedman and Wilcoxon test to better evaluate the performance of the algorithms for each database. First, we perform the Friedman test, if it returns $p_{value} \le 0.05$, it means that significant differences were found and we can proceed with a post-hoc procedure to characterize these differences. Next, we perform the Wilcoxon test for pairwise comparisons. If the Wilcoxon test returns $p_{value} \le 0.01$, it means that there is a significant difference between the compared pair \citep{derrac2011}. We applied these tests to all databases together and individually. For \textit{credit Australia} and \textit{spam}, the Friedman test returned $p = 0.841$ and $p = 0.180$, respectively. Thus, for these databases, there is no significant differences among the algorithms. Since the algorithms accuracy performance is too close, this is expected. For the rest of the databases, the values returned by the Friedman test is always less than 0.01, then we perform the Wilcoxon test for all of them. As we have too many pairwise comparisons, we decide to highlight the main test outcomes as follows:

\begin{itemize}
	\item For the \textit{alphabet}, \textit{DNA}, and \textit{madelon}, the RBM-ELM is significantly different when compared to all others. 
	
	\item For the \textit{gisette}, \textit{isolet} and \textit{urban cover land}, there is no significant difference between the ELM-AE and RBM-ELM. Moreover, they are significantly different when compared to all others.
	
	\item For the \textit{diabetic}, there is no significant difference between the ELM-AE and RBM-ELM. In addition, this pair is significantly different when compared to all others.
	
	\item In the \textit{overall}, the RBM-ELM is significantly different when compared to all others. 
\end{itemize}

\noindent According to the statistical test and the accuracy performance described in Table \ref{tab:results}, we conclude that for \textit{alphabet}, \textit{DNA}, and \textit{madelon}, the RBM-ELM is the best algorithm; For the \textit{gisette}, \textit{isolet} and \textit{urban cover land}, both RBM-ELM and ELM-AE are the best approaches; And for the \textit{diabetic}, the RBM-ELM and the ELM-RO are the best algorithms. Considering the overall result, for this group of benchmarks, our analysis indicates that the RBM-ELM is the best algorithm.

\subsection{Experiments remarks}
As we can see in sections \ref{sec:case-study} and \ref{sec:benchs}, the RBM-ELM improved the standard ELM and has better performance than the ELM-AE and ELM-RO. Nonetheless, our approach presents the highest time-consuming. Indeed, it is a drawback, however, for the gisette database, the largest one in our experiment, the RBM-ELM spent on average 160.23 seconds to improve almost 4\%, when compared to ELM. We consider it acceptable, since we have a good improvement on the final performance. Another issue about the RBM-ELM is the RBM parameters configuration. For some databases, we need to spend some time to find a good configuration. Unfortunately, to use the RBM we have to handle with this issue. Nonetheless, \cite{hinton2010} described a guide to setup it. Following this guide, we can reach good values for the RBM parameters in a faster way.

Although the statistical test points out that the RBM-ELM as the best algorithm for these databases, the ELM-AE and ELM-RO are still good approaches to improve the ELM. However, the RBM-ELM is more robust, since this approach is always in the group of the best algorithms for all databases. The ELM-AE and ELM-RO sometimes got bad performances such as in diabetic and alphabet databases, respectively. This does not occur with the proposed approach.   

\section{Conclusion} \label{sec:conclusion}
In this paper, we propose a new approach to determine the input weights for extreme learning machines (ELM) using the restricted Boltzmann machine (RBM), which we call RBM-ELM. In order to evaluate our new approach, we present an illustrative example detailing the RBM-ELM parameters configuration. Next, we carried out an experiment with standard benchmarks and compare the RBM-ELM performance with standard ELM, ELM autoencoder (ELM-AE) and a state of the art ELM random orthogonal (ELM-RO). The analysis of the results showed that the RBM-ELM was the best algorithm for the performed experiment and it was more stable than the other ones. On the other hand, our approach had the highest time-consuming among all algorithms. As we investigated, in this case it is worth to mention the trade-off between the improved accuracy and larger computational cost. In the future, we will work to improve the selection of the number of hidden neurons in the proposed approach.\\

\subsubsection*{Acknowledgments}
A. Pacheco and C. da Silva would like to thank the financial support of the Brazilian agency CAPES and R. Krohling thanks the financial support of the Brazilian agency CNPq under grant nr. 309161/2015-0.

\section*{References}
\bibliography{mybibfile}

\end{document}